\pdfoutput=1

\documentclass[11pt]{article}

\usepackage[]{ARR}

\usepackage{times}
\usepackage{latexsym}
\usepackage{amsmath}
\usepackage{array}
\usepackage{booktabs}
\usepackage{multirow}
\usepackage{amssymb}
\usepackage{graphicx}
\usepackage{enumerate}
\usepackage{diagbox}
\usepackage{microtype}
\usepackage{algorithm}
\usepackage{algpseudocode}

\usepackage{color}

\usepackage{cleveref}
\crefformat{section}{\S#2#1#3}
\crefformat{subsection}{\S#2#1#3}
\crefformat{subsubsection}{\S#2#1#3}
\crefrangeformat{section}{\S#3#1#4 to~\S#5#2#6}
\crefmultiformat{section}{\S#2#1#3}{ and~\S#2#1#3}{, #2#1#3}{ and~#2#1#3}
\Crefformat{figure}{#2Fig.~#1#3}
\Crefmultiformat{figure}{Figs.~#2#1#3}{ and~#2#1#3}{, #2#1#3}{ and~#2#1#3}
\Crefformat{table}{#2Tab.~#1#3}
\Crefmultiformat{table}{Tabs.~#2#1#3}{ and~#2#1#3}{, #2#1#3}{ and~#2#1#3}
\Crefformat{appendix}{#2Appx.~\S#1#3}

\usepackage[T1]{fontenc}

\usepackage[utf8]{inputenc}

\usepackage{microtype}

\usepackage{inconsolata}

\usepackage{amsmath}
\usepackage{amssymb}

\title{On-the-fly Denoising for Data Augmentation \\in Natural Language Understanding}

\author{Tianqing Fang$^{1}$\thanks{\quad Work done when visiting USC.}~, Wenxuan Zhou$^{2}$, Fangyu Liu$^3$, Hongming Zhang$^{4}$,\\
\textbf{Yangqiu Song$^{1}$, Muhao Chen$^{2,5}$}\\
$^{1}$Hong Kong University of Science and Technology~~~$^{2}$University of Southern California\\
$^{3}$University of Cambridge~~~$^{4}$Tencent AI Lab, Seattle~~~$^{5}$University of California, Davis\\
 \texttt{\{tfangaa, yqsong\}@cse.ust.hk}, \texttt{zhouwen@usc.edu}, \texttt{fl399@cam.ac.uk},\\ \texttt{hongmzhang@global.tencent.com}, \texttt{muhchen@ucdavis.edu}
\\
}

\begin{document}
\maketitle
\begin{abstract}
Data Augmentation (DA) is frequently used to provide additional training data without extra human annotation automatically.
However, data augmentation may introduce noisy data that impairs training.
To guarantee the quality of augmented data,
existing methods either assume no noise exists in the augmented data and adopt consistency training or use simple heuristics such as training loss and diversity constraints to filter out ``noisy'' data.
However, those filtered examples may still contain useful information, and dropping them completely causes a loss of supervision signals.
In this paper, based on the assumption that the original dataset is cleaner than the augmented data, we propose an on-the-fly denoising technique for data augmentation that learns from soft augmented labels provided by an organic teacher model trained on the cleaner original data.
To further prevent overfitting on noisy labels, a simple self-regularization module is applied to force the model prediction to be consistent across two distinct dropouts.
Our method can be applied to general augmentation techniques and consistently improve the performance on both text classification and question-answering tasks\footnote{Our code is available at \url{https://github.com/luka-group/ODDA-Data-Augmentation}}.
\end{abstract}

\section{Introduction}

The development of natural language understanding (NLU) comes along with the efforts in curating large-scale human-annotated datasets~\cite{DBLP:conf/nips/BrownMRSKDNSSAA20GPT3, DBLP:journals/corr/abs-2206-04615-imitation-game}.
The performance of NLP models usually highly correlates with the quantity and quality of training data.
However, human data annotations are usually expensive to acquire and hard to scale \cite{paulheim2018much}. To address this challenge, automatic data augmentation becomes an attractive approach to
effectively increase the scale of training data, and improve the performance of neural models, particularly in low-resource scenarios~\cite{DBLP:conf/emnlp/WeiZ19EDA, DBLP:conf/nips/XieDHL020UDA,DBLP:conf/emnlp/YangMFSBWBCD20GDAUG,DBLP:conf/acl/FengGWCVMH21}.

\begin{figure}[t]
    \centering
    \includegraphics[width=1.0\linewidth]{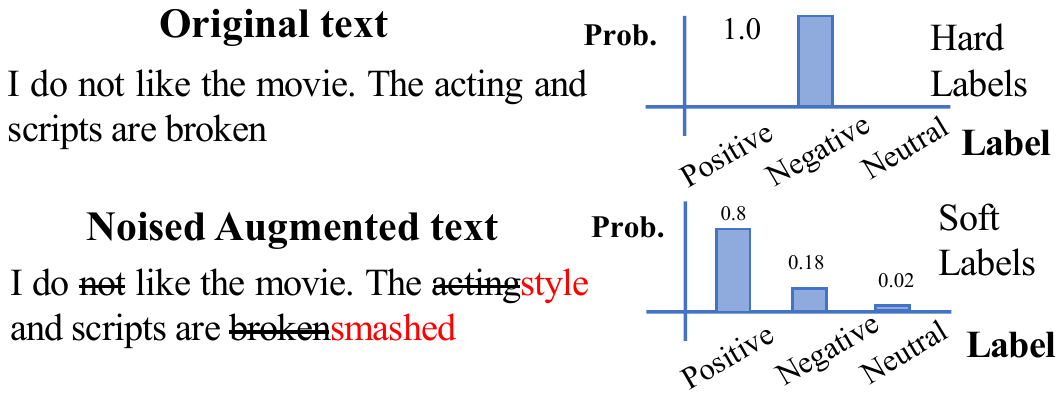}
    \caption{An example in a sentiment classification task about the noise brought by text-editing data augmentation.
    The noisy augmented text has the probability of being a ``positive'' attitude due to the removal of ``not''.
    }
    \label{fig:intro}
\end{figure}

However, automatic data augmentation techniques, regardless of token-level~\cite{DBLP:conf/emnlp/WeiZ19EDA,DBLP:conf/nips/XieDHL020UDA} or sentence-level~\cite{DBLP:conf/acl/SennrichHB16back-trans,DBLP:conf/emnlp/YangMFSBWBCD20GDAUG} ones,
may introduce noise to the augmented data. 
For example, in text classification or sentiment analysis tasks, altering or removing some decisive words can change the original label~\cite{DBLP:conf/coling/TroianoKP20}.
In addition, automatic data augmentation may distort the core semantic meaning or impair the fluency of the original text, leading to meaningless data instances~\cite{bayer2021survey}. 

To improve the quality of augmented data, various filtering techniques have been developed to select a subset of high-quality data.
Typical filtering paradigms design an uncertainty- or diversity-based metric to select data examples, for which the metric could be the loss of the task model trained on the original data~
\cite{DBLP:conf/naacl/ZhaoZ0DGZ22EPiDA, DBLP:conf/acl/KamallooR022Glitter}, diversity of the augmented data~\cite{DBLP:conf/naacl/ZhaoZ0DGZ22EPiDA, DBLP:conf/emnlp/YangMFSBWBCD20GDAUG, DBLP:conf/iclr/KimKAS22}, influence functions~\cite{DBLP:conf/emnlp/YangMFSBWBCD20GDAUG}, and 
logit consistency across multiple trained models~\cite{DBLP:conf/iclr/LiSH20, DBLP:conf/iclr/ZhouWB21}. 
However, data filtering mechanisms set a \textit{discrete} threshold and potentially discard examples that the model can still acquire signals from using properly designed denoising objectives~\cite{DBLP:conf/iclr/LiSH20}.
Alternative solutions to \textit{continuously} re-weighting~\cite{DBLP:conf/iclr/YiHSJLM21} augmented data or adopting consistency training~\cite{DBLP:conf/nips/XieDHL020UDA} often focus solely on the learnability of data or assume noisy examples should have the same label as the original ones, rather than mitigating their noise.



In this paper, we address the problem of \emph{learning from noisy augmented data} without 
(1) the effort of producing extra augmentations for filtering
and (2) the risk of losing useful supervision signals from examples that are \textit{discretely} filtered out.
Noisy data augmentation does not necessarily lead to a hard flipped label but a soft change in the original label distribution, as illustrated in \Cref{fig:intro}.
Therefore, we propose a soft noisy label correction framework called On-the-fly Denoising for Data Augmentation (ODDA), which distills task signals to noisy augmented instances and proactively mitigates noise.
Different from the \emph{learning from noisy label} (LNL) setting in 
fully supervised~\cite{wang-etal-2019-learning-noisy,wang-etal-2019-crossweigh,DBLP:conf/emnlp/ZhouC21_NLL} or distantly supervised training~\cite{DBLP:conf/emnlp/0001ZHWZJ021}, in data augmentation, the original dataset is cleaner and offers a natural distributional prior for estimating the noise level of augmented data, 
since the purpose of training data creation always involves approximating the data distribution in test time.
This assumption is also used in other works such as NoisyStudent~\cite{DBLP:conf/cvpr/XieLHL20}.
To leverage such signals, we propose an Organic Distillation\footnote{We call it \textit{organic} as the teacher model for distillation is trained on the original dataset.} module that uses a teacher model finetuned on the cleaner original dataset to provide soft labels for augmented data, where noisy data are softly relabeled to prevent the student model from overfitting to wrong labels.
Besides augmentation noise, the original data and organic distillation may also bring the noise. 
To address this issue, we further add a dropout-enabled self-regularization objective to force the predicted label distributions to be similar across two different dropout masks.
It is based on the observations that noisy labels may be forgotten during training or by perturbations, and self-regularization will force the consistency between perturbations and improve noise robustness~\cite{DBLP:conf/iclr/AghajanyanSGGZG21}. 

To summarize, the contributions of this paper are three-fold. 
First, we cast light on the problem of learning from noisy augmented data with \emph{soft label correction} instead of discretely filtering them out.
Second, we propose a simple yet effective on-the-fly denoising technique that 
continuously distills useful task signals to noisy augmentations,
coupled with a self-regularization loss to reduce overfitting to noise in general.
Third, we conduct extensive experiments on two NLU tasks, text classification and question answering, and show the effectiveness of our method for denoising both representative token-level and sentence-level data augmentation techniques.

\section{Related Works}

\paragraph{Data Augmentation and Filtering}

Recent studies on data augmentation for NLP have led to two main paradigms: \emph{token-level augmentation} and \emph{sentence-level augmentation}~\cite{DBLP:journals/corr/abs-2106-07499}.
Token-level augmentation conduct text editing on tokens from the input text.
Such techniques include using synonym replacement~\cite{DBLP:conf/nips/ZhangZL15, DBLP:conf/emnlp/WangY15,DBLP:conf/naacl/Kobayashi18} and word replacement with contextualized embedding or a masked language model~\cite{DBLP:conf/iclr/YiHSJLM21, DBLP:journals/corr/abs-2003-02245}, etc.
Particularly, EDA~\cite{DBLP:conf/emnlp/WeiZ19EDA} combines paraphrasing and random deletion, insertion, and swapping to perturb the text for augmentation. 
Sentence-level augmentation, on the other hand, modifies the whole sentence at once.
Methods include paraphrase-based augmentation techniques such as back-translation~\cite{DBLP:conf/acl/SennrichHB16back-trans, DBLP:conf/iclr/YuDLZ00L18} and paraphrase generation~\cite{DBLP:conf/coling/PrakashHLDQLF16}.
Another popular approach is to use conditional text generation models finetuned on the  
task dataset to automatically synthesize more training data.
It has been applied to tasks such as text classification~\cite{DBLP:conf/aaai/Anaby-TavorCGKK20, DBLP:journals/corr/abs-2003-02245}, machine reading comprehension~\cite{DBLP:conf/emnlp/PuriSSPC20}
, relation extraction~\cite{hu-etal-2023-gda}, 
commonsense reasoning~\cite{DBLP:conf/naacl/WestBHHJBLWC22, DBLP:conf/emnlp/YangMFSBWBCD20GDAUG}, and dialogue systems~\cite{kim2022soda}.
Another line of research operates on the embedding space. 
MIXUP-related augmentation generates augmented samples based on interpolating word embedding and label embedding vectors~\cite{DBLP:conf/acl/ChenYY20, DBLP:conf/acl/SiZQLWLS21}. 
Instead of focusing on concrete augmentation techniques, our paper study denoising synthetic data provided by any data augmentation method.

\begin{figure*}[t]
    \centering
    \includegraphics[width=1\linewidth]{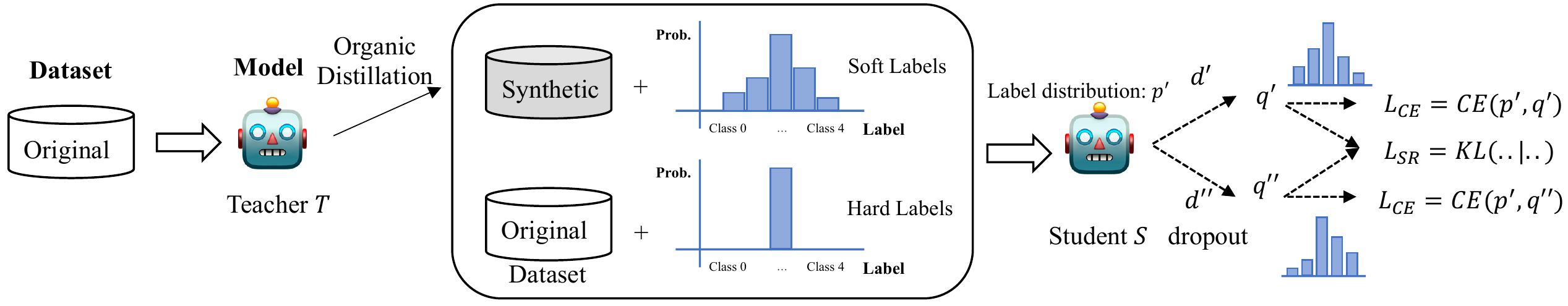}
    \caption{Overview of our ODDA framework.}
    \label{fig:overview}
\end{figure*}




\paragraph{Learning with Noisy Labels} 
In the field of NLP, particularly in low-resource settings, it is necessary to address the challenge of handling noisy labels derived from inaccurate annotations~\cite{DBLP:conf/emnlp/ZhouC21_NLL}, pseudo labels~\cite{DBLP:conf/iclr/LiSH20}, weak labels~\cite{DBLP:conf/naacl/ZengNFLZS22}, augmented data~\cite{DBLP:conf/acl/KamallooR022Glitter}, and other sources.
Various techniques have been developed to combat labeling noise in NLP datasets. 
Filtering-based techniques identify noisy examples through training dynamics or latent space features and then filter them out to produce a cleaner and more selective training dataset.
Such techniques are based on prediction consistency of different models~\cite{DBLP:conf/iclr/ZhouWB21}, loss-based uncertainty estimation~\cite{DBLP:conf/nips/HanYYNXHTS18}, and feature or representation-based outlier detection~\cite{DBLP:conf/nips/WuZ0M020,DBLP:conf/acl/FengGWCVMH21,DBLP:conf/cvpr/Wang0F22}. 
Besides 
noise filtering, an alternative approach to learning from noisy labels is to 
add an auxiliary learning objective to improve the noise robustness of a supervised model.
Techniques of this kind include
mixing up noisy examples~\cite{DBLP:conf/iclr/ZhangCDL18}, consistency training~\cite{DBLP:conf/nips/XieDHL020UDA, DBLP:conf/cvpr/XieLHL20}, co-regularization~\cite{DBLP:conf/emnlp/ZhouC21_NLL}, curriculum loss~\cite{DBLP:conf/iclr/LyuT20}, and semi-supervised training on noisy data~\cite{DBLP:conf/iclr/LiSH20}.

In data augmentation, recent studies have suggested using a filtering mechanism to select high-quality synthetic data from potentially noisy ones.
Typical filters include diversity~\cite{DBLP:conf/naacl/ZhaoZ0DGZ22EPiDA}, task loss~\cite{DBLP:journals/corr/abs-2210-07988}, consistency between two models~\cite{DBLP:conf/acl/0003XSHTGJ22}, influence function~\cite{DBLP:conf/emnlp/YangMFSBWBCD20GDAUG}, similarity with original data~\cite{avigdor-etal-2023-consistent}, and the alignment of the fully augmented Jacobian with labels/residuals~\cite{DBLP:journals/corr/abs-2210-08363}.
Instead of filtering, our method continuously learns from noisy labels with a cleaner teacher model and a denoising objective without discarding noisy instances, thus 
can more sufficiently acquire supervision signals from all augmented instances.
Our work also differs from consistency training, which assumes that augmented data, even if noisy, should have similar predictions to the original instances. In contrast, we aim to mitigate such noise, which runs counter to the objective of consistency training.


\section{Method}\label{sec:method}

This section introduces the problem formulation (\Cref{sec:problem_formulation}) and our ODDA framework (\Cref{sec:framework}-\Cref{ssec:training}).

\subsection{Problem Formulation}\label{sec:problem_formulation}

We consider the problem formulation of general text classification tasks.
We denote the dataset as $\mathcal{D}=\{(x_i, y_i)\}, i=1, \cdots, n$, where $x_i$ is the input text, $y_i\in \mathcal{Y}$ is the label of $x_i$ from the pre-defined label set $\mathcal{Y}$, and $n$ is the number of instances in the dataset. 
A data augmentation algorithm derives an augmented dataset $\mathcal{D}'=\{(x'_i, y'_i)\}, i=1, \cdots, kn$ from the original dataset $\mathcal{D}$, with an amplification factor $k$ denoting that for each data instance we generate $k$ augmentations.
We use both the original dataset $\mathcal{D}$ and the augmented dataset $\mathcal{D}'$ to train the classifier.
Other NLU tasks, such as sentiment analysis, multiple-choice question answering, and natural language inference, can be easily converted to a text classification paradigm.
For example, multiple-choice question answering can be converted to text classification by treating each question-answer pair as an input instance.

\subsection{On-the-fly Denoising}\label{sec:framework}

This subsection introduces the details of our On-the-fly Denoising for Data Augmentation (ODDA) framework. 
ODDA first trains an (organic) teacher model on the original dataset and then uses this teacher model to assign soft labels to the augmented dataset. 
During the learning process of augmented data, the model is jointly trained with two denoising objectives, where one is a cross-entropy loss on the distilled soft labels, and the other is a self-regularization loss to encourage robustness and consistency across two different dropout masks to automatically correct the noisy labels. 
The latter is important as the teacher model may also bring the noise to the soft labels, and self-regularization can serve as a general denoising channel for both forms of noise.
An overview illustration of ODDA is shown in \Cref{fig:overview}.

\paragraph{Organic Distillation (OD).} 
The first component of our framework 
is Organic Distillation.
We first train a teacher model on the original training dataset $D$. The resulting model (the \emph{organic teacher}), denoted as $T$, 
uses the same model architecture as the later student model.
Denote $z = f_T(x)$ as the function that produces logits $z$ given input $x$ using the teacher model $T$. 
For an instance $x$, the teacher model can predict the soft probability over the label set $\mathcal{Y}$ with a temperature-controlled softmax $g(z, \tau)$:

\begin{equation}\label{eq:temperature_softmax}
    q_y = g(z, \tau)_y = \frac{\exp{(z_y/\tau)}}{\sum_{j\in \mathcal{Y}} \exp{(z_j/\tau)}},
\end{equation}

\noindent
where $q_y$ is a predicted probability of a class $y$ from $\mathcal{Y}$, $\tau$ is a temperature hyperparameter where a larger temperature results in a smoother distribution.
Specifically, we omit $\tau=1$ in $g(\cdot, \tau)$, and use $g(x)$ to represent the standard softmax function.
We denote $f(x)$ as the student model that produces logits, and the loss function as cross-entropy loss $l_{\text{CE}}(p, q) = -(q \log p + (1-q) \log (1-p))$, where $p$ denotes the ground labels and $q$ denotes the predicted probabilities. 





Organic distillation distills knowledge from the organic teacher model 
to the augmented data.
As the original dataset is inherently of better quality than the augmented data, 
it can be used to provide a distributional prior on the level of noisiness in augmented data, thus calibrating the learning process of data augmentation and preventing overfitting 
the labeling noise.
For an augmented data instance $(x', y')$, we first compute the soft probabilities predicted by the organic teacher as $q'=g(f_T(x'), \tau)$, as in equation~(\ref{eq:temperature_softmax}).
Then $p'=g(f(x'))$ is the probability distribution over the label set $\mathcal{Y}$ predicted by the student model when training on synthetic data.
Then the corresponding loss function of organic distillation on the augmented example $x'$ is:
\vspace{-1em}

\begin{align}
    \mathcal{L}_{\text{OD}}(x') = & l_{\text{CE}}(p', q') \nonumber \\
    = & l_{\text{CE}}\Big(g\big(f(x')\big), g\big(f_T(x'), \tau\big)\Big)  .
\end{align}
\vspace{-1em}

\begin{algorithm}[t]
\small
\caption{On-the-fly DA Denoising (ODDA)}\label{alg:alg}
\begin{tabular}{p{2em}p{20em}}
\textbf{Input:} & Teacher model $f_T(\cdot)$, student model $f(\cdot)$, original dataset $\mathcal{D}=\{(x_i, y_i)\}, i=1,\cdots,n$, augmented dataset $\mathcal{D}'=\{(x_i', y_i')\}, i=1,\cdots,kn$, OD temperature $\tau$, SR coefficient $\alpha$. Max training steps for the organic teacher $s_T$ and the student $s_S$. \\
\textbf{Output:} & \hspace{0.5em} The trained student model $f(\cdot)$  \\
\end{tabular}

\begin{algorithmic}[1]
\State Initialize the teacher model $f_T(\cdot)$
\State $s \gets 0$ \Comment{Training steps for OD}
\While{$s < s_T$}
\State Sample a batch $\mathcal{B}$ from $\{(x_i, y_i)\}$
\State Train $f_T(\cdot)$ with cross-entropy loss on $\mathcal{B}$
\EndWhile
\State $s \gets 0$ \Comment{Training steps for Denoising}
\State $\mathcal{D}^+ \gets \{(x_i, y_i)\} \cup \{(x_i', y_i')\}$ \Comment{Mix $\mathcal{D}$ \& $\mathcal{D}'$}
\While{$s < s_S$}
\State Sample a batch $\mathcal{B}'$ from $\mathcal{D}^+$
\State Train $f(\cdot)$ with loss in Eq.~(\ref{eq:overall_loss}) on $\mathcal{B}'$ with Organic Distillation and Self-Regularization to do deonising
\EndWhile
\end{algorithmic}
\end{algorithm}

\paragraph{Self-Regularization (SR).} 
As the OD module may also introduce noise to the learning process, we introduce another general denoising channel.
Recent studies have shown that noisy instances generally tend not to be ``memorized'' easily by machine learning models, and are frequently ``forgetten'' given small perturbations~\cite{DBLP:conf/nips/XieDHL020UDA, DBLP:conf/iclr/AghajanyanSGGZG21} and along with the training steps~\cite{DBLP:conf/emnlp/ZhouC21_NLL}.
The often inconsistent characteristics of noisy instances over the learning curve is mainly attributed to their contradiction to the model's overall task inductive bias represented coherently by the clean data. 
To mitigate the impact of noise from individual data instances, inconsistent outputs resulting from small perturbations should be corrected."
Instead of filtering noisy examples out with the risk of losing useful information, we learn from noisy (and clean) examples with an additional objective by bounding the model's output to be consistent under small perturbations. 
Following R-Drop~\cite{DBLP:conf/nips/LiangWLWMQCZL21}, the perturbations are introduced with dropout,
and a regularization loss forcing the model prediction to be consistent across two different dropout outputs is adopted\footnote{A detailed explanation and theoretical analysis to self-regularization is presented in \Cref{sec:appendix_sr}.}.
Denote $d(f(x))$ as the function that outputs the predicted probability distribution under a dropout mask $d$, and $d_i$ is the $i$-th dropout mask. 
Then the self-regularization loss is defined as the Kullback-Leibler (KL) divergence between the average probability distribution of the $m$ dropout operations and the output of each dropout:

\begin{align}
    \bar{p} &= \frac{1}{m} \sum_{i=1}^m  g(d_i(f(x'))) , \nonumber \\%
    \mathcal{L}_{\text{SR}}(x') &= \frac{1}{m} \sum_{i=1}^m \text{KL}\Big( \bar{p} || g\big(d_i(f(x'))\big) \Big)  .
\end{align}

\subsection{Joint Training}\label{ssec:training}  
In the end, the model is jointly trained with the \textbf{OD} and \textbf{SR} objectives on the original dataset $\{(x_i, y_i)\}$ and the augmented dataset $\{(x_i', y_i')\}$:


\begin{align}    
    \mathcal{L} = & \frac{1}{n}\sum_{i=1}^n l_{\text{CE}}\Big(g\big(f(x_i)\big), y_i\Big) \nonumber \\ 
    & + \frac{1}{kn}\sum_{i=1}^{kn} \mathcal{L}_{\text{OD}}(x_i') \nonumber \\
    & + \alpha \frac{1}{kn+n} \sum_{i=1}^{kn+n}  \mathcal{L}_{\text{SR}}(x_i')\label{eq:overall_loss} .
\end{align}

The overall loss function is the sum of the cross-entropy loss on the original data with hard labels, the cross-entropy loss of the augmented data with soft labels distilled with the organic teacher, and the KL divergence between the average probability across $m$ different dropouts and each of the $m$ dropouts.
Here $l_{\text{CE}}( \cdot )$ is the cross-entropy loss function, 
$n$ is the number of original examples and $k$ is the amplification factor for data augmentation,
and $\alpha$ is a hyper-parameter to control the effect of self-regularization. 
In the third term, the SR is applied to both the original and augmented data, where the number of instances $n+kn$ indicates the collection of both the original and augmented data.
Though we derive these formulations based on the text classification task, in multiple-choice QA tasks,
the formulation can be accordingly converted to a $c$-class classification task, where $c$ is the number of choices per question.
The algorithm is outlined in Alg. \ref{alg:alg}.

\begin{table*}[t]
\small
\centering
\renewcommand\arraystretch{1.1}
\setlength\tabcolsep{1.9pt}
\begin{tabular}{lcccccccccc}
\toprule
\multirow{2}{*}{Method} & \multicolumn{2}{c}{TREC} & \multicolumn{2}{c}{Irony} & \multicolumn{2}{c}{AGNews} & \multicolumn{2}{c}{Sentiment} & \multicolumn{2}{c}{Offense} \\ \cline{2-11}
& 1\% & 10\% & 1\% & 10\% & 0.05\% & 0.1\% & 1\% & 10\% & 0.1\% & 1\% \\
\midrule
Sup. & 60.64$_{\pm\text{0.60}}$ & 90.53$_{\pm\text{0.47}}$ & 55.48$_{\pm\text{1.05}}$ & 63.14$_{\pm\text{0.99}}$ & 84.05$_{\pm\text{0.47}}$ & 86.43$_{\pm\text{0.07}}$ & 54.10$_{\pm\text{1.22}}$ & 65.56$_{\pm\text{0.22}}$ & 51.91$_{\pm\text{0.53}}$ & 64.35$_{\pm\text{0.12}}$\\
\midrule
\multicolumn{11}{c}{\textbf{Data Augmentation}} \\
EDA & 61.68$_{\pm\text{0.29}}$ & 93.83$_{\pm\text{0.63}}$ & 57.07$_{\pm\text{0.66}}$ & 64.55$_{\pm\text{0.52}}$ & 84.01$_{\pm\text{0.18}}$ & 86.43$_{\pm\text{0.07}}$ & 56.57$_{\pm\text{0.75}}$ & 65.80$_{\pm\text{0.14}}$ & 51.86$_{\pm\text{0.37}}$ & 64.61$_{\pm\text{0.15}}$ \\
EPiDA & 64.92$_{\pm\text{0.50}}$ & 93.96$_{\pm\text{0.18}}$ & 58.25$_{\pm\text{0.95}}$ & 64.72$_{\pm\text{0.58}}$& 84.51$_{\pm\text{0.31}}$& 86.68$_{\pm\text{0.19}}$& 57.20$_{\pm\text{0.32}}$& 65.58$_{\pm\text{0.24}}$& 51.55$_{\pm\text{0.49}}$ & 64.45$_{\pm\text{0.16}}$ \\
Glitter& 64.16$_{\pm\text{0.20}}$ & 93.55$_{\pm\text{0.06}}$ & 58.76$_{\pm\text{0.44}}$ & 64.73$_{\pm\text{0.95}}$ & 84.84$_{\pm\text{0.32}}$ & 87.00$_{\pm\text{0.29}}$ & \textbf{57.73}$_{\pm\text{0.31}}$ & 65.52$_{\pm\text{0.20}}$ & 51.69$_{\pm\text{0.42}}$ & 64.45$_{\pm\text{0.15}}$ \\
Large-loss & 62.21$_{\pm\text{1.71}}$ & 94.06$_{\pm\text{1.90}}$ & 57.07$_{\pm\text{2.13}}$ & 64.42$_{\pm\text{1.28}}$ & 83.48$_{\pm\text{0.97}}$ & 86.43$_{\pm\text{0.28}}$ & 57.13$_{\pm\text{1.27}}$ & 65.66$_{\pm\text{0.49}}$ & 51.78$_{\pm\text{0.77}}$ & 64.49$_{\pm\text{0.41}}$ \\
Re-weight & 64.37$_{\pm\text{1.69}}$ & 95.28$_{\pm\text{0.97}}$ & 58.14$_{\pm\text{2.34}}$ & 64.56$_{\pm\text{1.73}}$ & 84.45$_{\pm\text{1.12}}$ & 86.82$_{\pm\text{0.50}}$  & 56.81$_{\pm\text{1.52}}$ & 65.55$_{\pm\text{1.50}}$ & 51.70$_{\pm\text{1.10}}$ & 64.54$_{\pm\text{0.43}}$ \\
Consist. & 65.55$_{\pm\text{0.81}}$ & 95.15$_{\pm\text{0.90}}$ & 58.32$_{\pm\text{1.71}}$ & 64.50$_{\pm\text{1.24}}$ & 84.34$_{\pm\text{0.78}}$ & 86.45$_{\pm\text{0.26}}$ & 57.10$_{\pm\text{1.26}}$ & 65.64$_{\pm\text{0.46}}$ & 51.86$_{\pm\text{0.98}}$ & 64.66$_{\pm\text{0.43}}$ \\
\midrule
\multicolumn{11}{c}{\textbf{Denoising Data Augmentation (EDA as the DA algorithm)}} \\
Ours (OD) & 65.17$_{\pm\text{1.25}}$ & 95.02$_{\pm\text{1.42}}$ & 58.51$_{\pm\text{2.67}}$ & 64.73$_{\pm\text{0.18}}$ & 84.91$_{\pm\text{0.44}}$ & 86.84$_{\pm\text{0.26}}$ & 57.09$_{\pm\text{1.63}}$ & 65.68$_{\pm\text{0.51}}$ & 52.13$_{\pm\text{1.43}}$ & 65.16$_{\pm\text{0.64}}$  \\
Ours (SR) & 65.87$_{\pm\text{1.22}}$ & 95.50$_{\pm\text{0.68}}$ & 57.51$_{\pm\text{1.92}}$ & 64.24$_{\pm\text{0.61}}$ & 84.80$_{\pm\text{0.57}}$ & 86.75$_{\pm\text{0.57}}$ & 57.42$_{\pm\text{1.09}}$ & 65.74$_{\pm\text{0.27}}$ & 52.01$_{\pm\text{0.99}}$ & 65.06$_{\pm\text{0.49}}$ \\
Ours (both) & \textbf{67.16}$_{\pm\text{0.37}}$ & \textbf{96.04}$_{\pm\text{0.08}}$ & \textbf{60.66}$_{\pm\text{1.43}}$ & \textbf{65.54}$_{\pm\text{0.37}}$ & \textbf{86.30}$_{\pm\text{0.13}}$ & \textbf{87.14}$_{\pm\text{0.17}}$ & 57.17$_{\pm\text{0.37}}$ & \textbf{65.90}$_{\pm\text{0.19}}$ & \textbf{52.34}$_{\pm\text{0.53}}$ & \textbf{65.43}$_{\pm\text{0.29}}$ \\
\bottomrule
\end{tabular}
\caption{Performance of different filtering and re-weighting methods on the five text classification datasets, where EDA is used as the base data augmentation algorithm for all methods. 
1\% means using 1\% of the original training data for training. 
We report the average f1 score across five different random seeds. 
}
\label{table:text_classification}
\end{table*}

\section{Experiments}\label{sec:exp}


This section introduces experimental settings and results analysis. 
We evaluate on two representative tasks in NLU, few-shot text classification (Section~\Cref{sec:exp_text_cls}) and multiple-choice (commonsense) question answering (Section~\Cref{sec:exp_csqa}).
We use EDA~\cite{DBLP:conf/emnlp/WeiZ19EDA} as a representative token-level based augmentation method for text classification, 
and use Generative Data Augmentation (G-DAUG)~\cite{DBLP:conf/emnlp/YangMFSBWBCD20GDAUG} to explore task-aware sentence-level augmentation methods for hard QA tasks that require commonsense reasoning abilities.
In Section~\Cref{sec:ablation}, we provide ablation studies to show the effect of ODDA under synthetic noise on augmented data, the influence of hyperparameters, and the effect of denoising modules. 

\subsection{Text Classification} \label{sec:exp_text_cls}


\paragraph{Setup.}
Following the previous work~\cite{DBLP:conf/naacl/ZhaoZ0DGZ22EPiDA}, we use five text classification datasets: \textbf{TREC}~\cite{DBLP:conf/coling/LiR02} (Question classification, $n$=5,452), \textbf{Irony}~\cite{DBLP:conf/semeval/HeeLH18} (Tweets Irony Classification, $n$=3,817), \textbf{AGNews}~\cite{DBLP:conf/nips/ZhangZL15} (News Classification, $n$=120,000), \textbf{Sentiment}~\cite{DBLP:conf/semeval/RosenthalFN17} (Tweets Sentiment Analysis, $n$=20,631), and \textbf{Offense}~\cite{DBLP:conf/icwsm/FountaDCLBSVSK18} (Tweets Offense Detection, $n$=99,603).
We randomly sample different proportions of each dataset for experiments to fully demonstrate the effect of data augmentation, where the percentage in \Cref{table:text_classification} (\%) indicates the percentage of data sampled for training,
leading to
around 100 and 1000 examples sampled for the two few-shot proportions, respectively.
BERT-base~\cite{DBLP:conf/naacl/DevlinCLT19} is used as the backbone model for all the text classification experiments, 
which is incorporated with EDA~\cite{DBLP:conf/emnlp/WeiZ19EDA} for data augmentation.
The augmentation probability of the four edit operations in EDA is equally set as 0.05.
We report the average macro-F1 across five different random seeds and the standard deviation in subscripts.
Each original data example is associated with $k=3$ augmented data.
The OD temperature $\tau$ is searched within \{0.5, 1, 2, 3\}, and the SR $\alpha$ is searched within \{5, 10, 20, 50, 100\}.
Early stopping is used to select the model with the best performance.
More hyperparameters are shown in~\Cref{sec:hparam_text_cls}.


\paragraph{Baselines.} 
We compare three types of baseline denoising techniques, which are filtering, re-weighting, and consistency training.
For filtering, we use EPiDA (Relative Entropy Maximization + Conditional Entropy Minimization, \citet{DBLP:conf/naacl/ZhaoZ0DGZ22EPiDA}), Glitter (selecting augmented data with higher task loss, \citet{DBLP:conf/acl/KamallooR022Glitter}), 
Large-loss (select augmented data with small loss, \citet{DBLP:conf/nips/HanYYNXHTS18}), 
to filter out low-quality augmented training data.
For re-weighting, we use the re-weighting factors in \citet{DBLP:conf/iclr/YiHSJLM21}, where examples with larger training loss are given larger weights.
For consistency training (denoted as Consist.), we use the idea in Unsupervised Data Augmentation (UDA; \citealp{DBLP:conf/nips/XieDHL020UDA}) to add a consistency loss between original examples and the corresponding augmented examples.
More details are provided in~\Cref{sec:hparam_text_cls}.

\paragraph{Results and Analysis.}
The main experimental results of text classification are presented in \Cref{table:text_classification}. 
First, we can see that ODDA can provide remarkable improvements over EDA, the base data augmentation method without any filtering or denoising. The notable improvement of F1 2.5\% increase in average for the smaller few-shot split and 1.0\% F1 increase in average for the larger few-shot split over EDA indicate the importance of addressing the noise issue in augmented data.

Second, ODDA outperforms filtering-based baselines (EPiDA, Glitter, and Large-loss) in all datasets and splits except for the 1\% Sentiment. Note that these baselines need to select $k=3$ augmented examples per original example from a candidate pool of $50$ EDA-generated augmented examples per original example, while in our method directly generates the $k=3$ augmented examples per original instance.
Those filtering baselines are more costly and require generating 16 times more augmentations than our method to perform filtering.
We can conclude that learning with a denoising objective for data augmentation can be far more data efficient than filtering by exploiting the denoising training signals from noisy examples without filtering them out.

Third, ODDA outperforms re-weighting and Consist. by a large margin. 
These two methods adopt an opposite idea of denoising to some extent. 
For re-weighting, augmented examples with larger training loss, which can be regarded as more noisy~\cite{DBLP:conf/nips/ShuXY0ZXM19}, will be up-weighted during training, while in our Organic Distillation and Sefl-regularization, examples identified noisier will be down-weighted to rectify the effect of noisy augmented instances.
For Consistency training, it assumes that the original and its corresponding augmented example should share the same label and train them with a consistency loss, which is also opposite to our assumption that augmented data may be noisy.
From the comparison of those two methods, we can conclude that the denoising objective better suits the scenario of data augmentation than both the learnability-based re-weighting and 
the consistency training with label-preserving assumption.

\begin{table*}[t]
\small
\centering
\renewcommand\arraystretch{1.1}
\setlength{\tabcolsep}{2mm}{} 
\begin{tabular}{l|cccccc|p{5em}<{\centering}}
\toprule
  & \multicolumn{6}{c|}{WinoGrande} & \multirow{2}{*}{CSQA}\\
  & XS & S & M & L & XL & AUC & \\
\midrule
Supervised & 60.28\tiny{$\pm$1.52} & 62.23\tiny{$\pm$2.06} & 66.00\tiny{$\pm$1.28} & 74.68\tiny{$\pm$0.28} & 79.09\tiny{$\pm$0.56} & 68.12 &76.35\tiny{$\pm$0.31} \\
G-DAUG & 60.49$_{\pm\text{0.44}}$ & 66.04$_{\pm\text{0.48}}$ & 72.22$_{\pm\text{0.43}}$ & 76.79$_{\pm\text{0.77}}$ & 80.09$_{\pm\text{0.53}}$ & 71.32 & 77.38$_{\pm\text{0.36}}$ \\
Ours (OD) & 61.18$_{\pm\text{0.59}}$ & 67.45$_{\pm\text{0.47}}$ & 72.38$_{\pm\text{0.73}}$ & 77.35$_{\pm\text{0.22}}$ & 80.75$_{\pm\text{0.36}}$ & 72.01 & 78.41$_{\pm\text{0.40}}$ \\
Ours (SR) & 60.68$_{\pm\text{0.72}}$ & 67.06$_{\pm\text{0.69}}$ & 72.34$_{\pm\text{0.68}}$ & 77.09$_{\pm\text{0.38}}$ & 80.57$_{\pm\text{0.56}}$ & 71.76 & 77.62$_{\pm\text{0.41}}$ \\
Ours (both) & \textbf{61.30}$_{\pm\text{0.55}}$ & \textbf{67.62}$_{\pm\text{0.48}}$ & \textbf{72.68}$_{\pm\text{0.70}}$ & \textbf{77.65}$_{\pm\text{0.21}}$ & \textbf{80.80}$_{\pm\text{0.51}}$ & \textbf{72.23} & \textbf{78.69}$_{\pm\text{0.31}}$ \\
\bottomrule
\end{tabular}
\caption{Performance of commonsense question answering.} \label{table:csqa}
\vspace{-1em}
\end{table*}

\subsection{Commonsense Question Answering} \label{sec:exp_csqa}

\paragraph{Setup.}
We follow the setups in G-DAUG~\cite{DBLP:conf/emnlp/YangMFSBWBCD20GDAUG} to conduct commonsense QA experiments.
We study a full-shot setting here for the QA tasks as a supplement to the few-shot text classification experiments, and select two representative multiple-choice commonsense QA datasets, WinoGrande~\cite{DBLP:conf/aaai/SakaguchiBBC20} and CommonsenseQA (CSQA; \citealt{DBLP:conf/naacl/TalmorHLB19}). 
Other datasets are not selected as they either adopt a few-shot setting, or the augmented data is not publicly available. 
We use the released version of augmented data by \citet{DBLP:conf/emnlp/YangMFSBWBCD20GDAUG}\footnote{https://github.com/yangyiben/G-DAUG-c-Generative-Data-Augmentation-for-Commonsense-Reasoning} produced with finetuned GPT-2~\cite{radford2019gpt2}.
RoBERTa-large~\cite{DBLP:journals/corr/abs-1907-11692} is used as the backbone QA model, and the hyperparameters are the same as in \citet{DBLP:conf/emnlp/YangMFSBWBCD20GDAUG}. 
We evaluate the model performance using accuracy for each subset in WinoGrande, and an AUC calculated with 
the curve of the logarithm of the number of instances of each subset against the corresponding accuracy, to present an overall performance on WinoGrande across the five subsets.
Accuracy is used for CSQA as the evaluation metric. 
As linear learning rate decay is applied during the training, we report the performance of the last checkpoint during training.
Different from the original paper of G-DAUG~\cite{DBLP:conf/emnlp/YangMFSBWBCD20GDAUG}, which reports the performance of only one run, we report the average and standard deviation across five different random seeds.
More details about models and datasets are presented in~\Cref{sec:hparam_qa}.

\begin{figure}[t]
    \centering
    \includegraphics[width=1\linewidth]{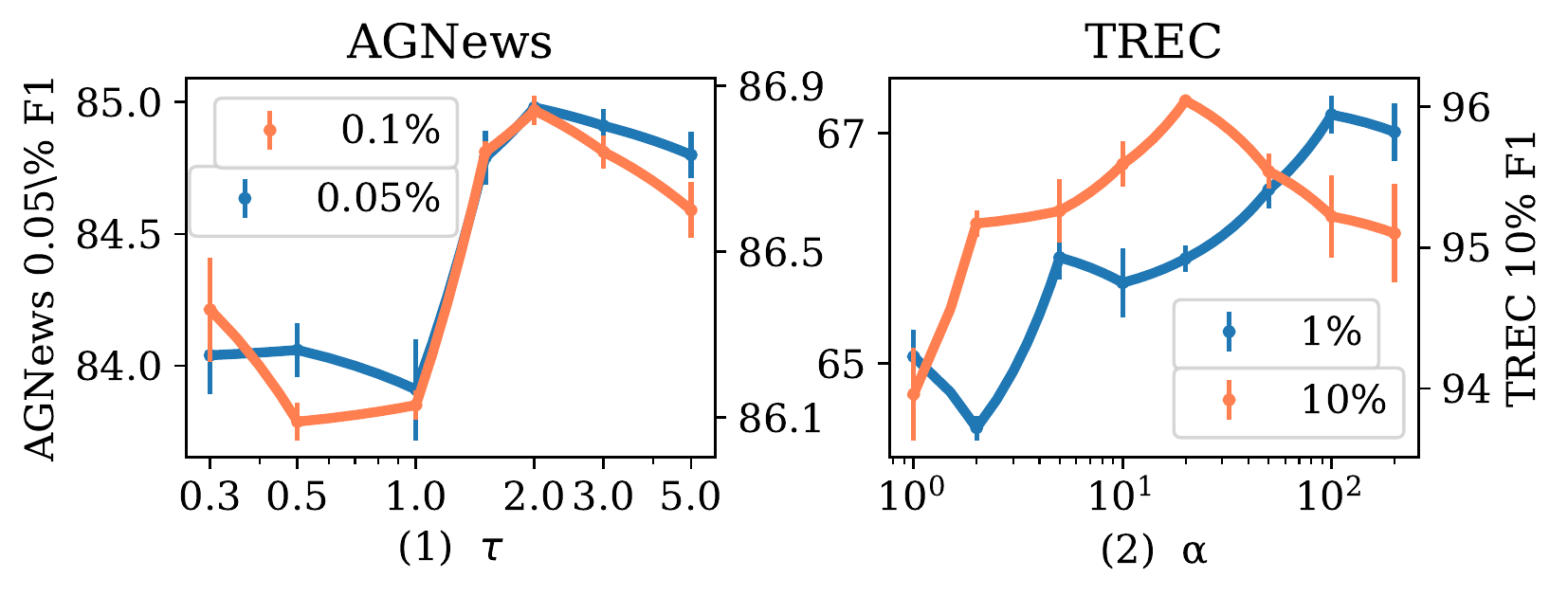}
    \caption{(1) The effect of OD temperature $\tau$ on the classification performance for AGNews dataset. (2) The effect of SR coefficient $\alpha$ on the classification performance for TREC dataset.}
    \label{fig:ablations}
\end{figure}

\paragraph{Baselines.}
As in G-DAUG, the augmented instances are already filtered with an influence function~\cite{DBLP:conf/icml/KohL17} and diversity heuristics, we do not conduct further filtering as baselines.
And as no 
direct mapping exists between the original and augmented examples, the re-weighting and consistency training baseline does not fit the sentence-level data augmentation setting.
Hence, we only compare the performance of adding our on-the-fly denoising technique on top of the already-filtered augmented dataset against the performance of G-DAUG and the supervised learning baseline without data augmentation. 
We also check the effect of each channel (OD and SR).


\paragraph{Results and Analysis.}
The QA results are shown in \Cref{table:csqa}. 
When we apply ODDA to the augmented data generated by G-DAUG filtered with influence function and a diversity heuristic defined in~\citet{DBLP:conf/emnlp/YangMFSBWBCD20GDAUG}, the performance can be consistently improved across different few-shot splits of WinoGrande and full-shot CSQA.
These experiments first demonstrate that besides token-level data augmentation, where each augmented example can be aligned with its original example, ODDA can also work well for sentence-level data augmentation, where there is no explicit mapping between augmented data and original data. 
This is an advantage as some data augmentation boosting methods need to leverage the mapping between original and augmented examples to select semantically similar augmentations (e.g., EPiDA) or use consistency training, while our method is not restricted 
by this precondition.
Second, we show that our method can not only be used for boosting text classification, but can work well for more complex commonsense reasoning tasks.






\subsection{Ablation Study}\label{sec:ablation}

\noindent \textbf{Organic teacher distillation.} The Organic Distillation (OD) module distills the knowledge from the relatively cleaner original dataset to the augmented data with soft labels, preventing overfitting on hard noisy labels.
We check the influence of the distillation temperature $\tau$ on the model performance, shown in \Cref{fig:ablations} (1) for the AGNews dataset as an example. 
Specifically, the model performance reaches its best when the temperature $\tau=2$, indicates a softer label distribution.
For other datasets such as TREC, Irony, and Offense, the variance of different temperatures is relatively minor, and we select $\tau=1$ as the default. 
While for AGNews and Sentiment, the model can benefit from larger temperature, which may indicate that there is more noise in the augmented data from those two datasets, and softer distribution help reduce overfitting on the augmented data.

\paragraph{Self-regularization.}
The self-regularization (SR) module in our framework serves as a general denoising channel to minimize the discrepancy of model outputs between two dropouts. The $\alpha$ in \Cref{eq:overall_loss} is the hyperparameter measuring the importance of the denoising effect.
We take the TREC dataset as an example to show the effect of $\alpha$ on the model performance as in \Cref{fig:ablations} (2).
We can see that for TREC 1\%, the performance reaches the maximum when $\alpha=100$, and for TREC 10\%, the model performs the best when $\alpha=20$. 
Such a difference indicates that in TREC 1\%, which contains only fewer than 100 training examples, it can benefit more when the effect of self-regularization out-weight the original cross-entropy loss. Similar results are shown in other datasets under the smaller few-shot training set.

\begin{table}[t]
\small
\centering
\renewcommand\arraystretch{1.1}
\setlength\tabcolsep{2.0pt}
\begin{tabular}{lccccc}
\toprule
\multirow{2}{*}{Method} &  \multicolumn{4}{c}{Irony 10\%} \\\cline{2-5}
& $p_n=0.0$ & $p_n=0.1$ & $p_n=0.3$ & $p_n=0.5$ \\ 
\midrule
EDA & 64.55 & 63.27 & 63.26 & 60.41  \\
EPiDA & 64.72 & 64.57 & 63.94 & 63.24 \\
Glitter & 64.73 & 65.04 & 62.99 & 61.85 \\
Large-loss & 64.42 & 63.42 & 63.27 & 61.56 \\
Re-weight & 64.56 & 64.38 & 64.53 & 63.79 \\
Ours (both) & \textbf{65.54} & \textbf{65.54} & \textbf{65.54} & \textbf{65.54} \\
\bottomrule
\end{tabular}
\caption{Experiments on adding synthetic noise to augmented data for the Irony dataset (10\%), when original data remain still. 
$p_n$ indicates the probability that the label of an augmented example is flipped.
As our method learns with the soft labels provided by the clean original dataset, it is not affected by noise on labels in the augmented dataset.}
\label{table:noise_study}
\end{table}

\paragraph{Adding synthetic noise.}
We further show the effect of our denoising method by introducing synthetic noise of different levels to augmented data. 
The original dataset remains unchanged to show the effect of a cleaner original dataset.
To better demonstrate the effect of denoising in augmented data, we control the noise level by setting a probability $p_n$ of flipping the label of augmented data.
We select the dataset Irony (with 10\% training data) as an example, as Irony is a binary classification task and flipping the label will definitely lead to an opposite label (for other datasets such as AGNews, there may be slight overlaps between different labels). 
The results are presented in \Cref{table:noise_study}. We can see that EDA and all filtering methods suffer from performance degradation along with increased noise proportions, while our method is not influenced by such synthetic noise as we do not rely on the hard label of augmented data but the soft label provided by the organic teacher model.
The performance degradation is not too drastic when $p_n$ increases as the labels of original data are retained.
Such an experiment further consolidates the effectiveness of our denoising method for data augmentation.

\begin{table}[t]
\small
\centering
\renewcommand\arraystretch{1.1}
\setlength\tabcolsep{2.0pt}
\begin{tabular}{lcccccc}
\toprule
\multirow{2}{*}{Method} & \multicolumn{2}{c}{TREC} & \multicolumn{2}{c}{Irony} & \multicolumn{2}{c}{AGNews} \\ \cline{2-7}
& 1\% & 10\% & 1\% & 10\% & 0.05\% & 0.1\%  \\
\midrule
Iter. Teacher & 66.89 & 95.56 & 58.73 & 64.49 & 84.15 & 86.17 \\
EMA & 64.10 & 95.26 & 57.37 & 64.40 & 84.16 & 86.36 \\
Co-Reg & 65.19 & 95.08 & 58.29 & 64.86 & 84.81 & 86.54  \\
Co-Teaching & 64.62 & 94.69 & 57.39 & 65.51 & 84.83 & 86.91 \\
Ours (SRx3) & 66.19 & 95.54 & 58.31 & 64.56 & 84.44 & 86.56  \\
Ours (SRx4) & 65.88 & 95.69 & 58.95 & 64.62 & 84.67 & 86.33 \\
\midrule 
Ours (OD) & 65.17 & 95.02 & 58.51 & 64.73 & 84.91 & 86.84  \\
Ours (SR) & 65.87 & 95.50 & 57.51 & 64.24 & 84.80 & 86.75 \\
Ours (both) & \textbf{67.16} & \textbf{96.04} & \textbf{60.66} & \textbf{65.54} & \textbf{86.30} & \textbf{87.14} \\
\bottomrule
\end{tabular}
\caption{Ablations on the effect of Organic Distillation (OD) and Self-Regularization (SR), compared to their counterparts. SRx$n$ means dropouts are done $n$ times.}
\label{table:ablation_component}
\end{table}

\paragraph{Alternative denoising techniques.}
We also study the alternative solutions to our denoising framework.
There are alternative ways to the organic teacher. For example, we could iteratively select the best-performed teacher model during the training with augmented data (denoted as an iterative teacher).
For the general denoising channel SR, there are other techniques that perform denoising, such as using Exponential Moving Average (EMA) over training steps~\cite{DBLP:conf/nips/TarvainenV17}, or using the consistency of two independently-trained models to perform logits regularization~\cite{DBLP:conf/emnlp/ZhouC21_NLL}.
We also study whether increasing the number of dropouts $m$ to do regularization will help the model performance.
These experiments are collectively presented in \Cref{table:ablation_component}.
We can see that our proposed method achieves the best among other alternative choices. For the Iterative Teacher, though the teacher model is iteratively updated, 
it may lose the information by cleaner original dataset when further trained on the augmented data.
For Co-Regularization, it achieves similar performance when two identical models are simultaneously trained to improve consistency. However, it doubles the cost of training.
When doing multiple dropouts in self-regularization, the performance on the 1\% split of TREC and Irony can be improved when $m>2$, while for others, the improvements are not significant. Considering that using $m=3 \text{ or } 4$ will lead to 1.5 and 2 times the computational cost, we choose $m=2$ to make the training more efficient while keeping competitive results.







\section{Conclusion}

In this paper, we address the problem of improving data augmentation via denoising, and propose an efficient on-the-fly data augmentation denoising framework that leverages a teacher model trained on the cleaner original dataset for soft label correction and a self-regularized denoising loss for general denoising. 
Such a denoising pipeline can well benefit the tasks with limited annotated data and noisy augmented data.
Experiments show that our denoising framework performs consistently better than the baselines of filtering, re-weighting, and consistency training, 
with both token-level and sentence-level data augmentation methods on few-shot text classification and commonsense question-answering tasks.

\section*{Acknowledgement}
Tianqing Fang was supported by the Hong Kong PhD Fellowship Scheme.
Wenxuan Zhou and Muhao Chen were supported by the NSF Grants IIS 2105329 and ITE 2333736, an Amazon Research Award and a Cisco Research Award.
Yangqiu Song was supported by the NSFC Fund (U20B2053) from the NSFC of China, the RIF (R6020-19 and R6021-20) and the GRF (16211520 and 16205322) from RGC of Hong Kong. 
Yangqiu Song thanks the support from the UGC Research Matching Grants (RMGS20EG01-D, RMGS20CR11, RMGS20CR12, RMGS20EG19, RMGS20EG21, RMGS23CR05, RMGS23EG08).
Tianqing Fang and Yangqiu Song also thank the support from Tencent AI lab.

\section*{Limitations}

We only include one representative token-level and sentence-level data augmentation technique in our experiments, while cannot enumerate all others such as masked language models replacing~\cite{DBLP:conf/iclr/YiHSJLM21}. 
In addition, we only include two representative NLU tasks in the experiments while others such as natural language inference~\cite{DBLP:conf/emnlp/BowmanAPM15} are missing due to the limited presentation space.
As for the method ODDA itself, we conduct denoising using the training information within a single training step without considering longer dependencies and training dynamics across different training steps or epochs, which can be a future work of this study.

\bibliography{custom}

\clearpage

\appendix

\begin{center}
    {
    \Large\textbf{Appendices}
    }
\end{center}

\section{More Details about Experiments}

\subsection{More Details about Text Classification} \label{sec:hparam_text_cls}

We use the codebase and experimental settings from EPiDA\footnote{https://github.com/zhaominyiz/EPiDA}~\cite{DBLP:conf/naacl/ZhaoZ0DGZ22EPiDA} to conduct our experiments. Table~\ref{table:hparam_text_cls} shows the essential hyperparameters that are used for each dataset. 
During the training, we first train a few epochs on the original dataset, and then finetune on the union of augmented data and original data. 

For EPiDA~\cite{DBLP:conf/naacl/ZhaoZ0DGZ22EPiDA}, we follow the setting in the original paper to first produce $k=50$ augmented examples per original example using EDA, and then select top 3 scored by its Relative Entropy Maximization (REM) and Conditional Entropy Minimization (CEM) filter. The trade-off parameter between REM and CEM is set as 0.5, as in the original paper. 

For Glitter~\cite{DBLP:conf/acl/KamallooR022Glitter} and large-loss, similar with EPiDA, we sample 50 augmented examples first, and select the top 3 examples with the largest/smallest loss in the current run.
For Re-weight~\cite{DBLP:conf/iclr/YiHSJLM21}, we use the following re-weighting equation to re-weight the augmented data in a batch:

$$w_{x_i}=\frac{ \exp{\Big( \frac{1}{\lambda} l_{\text{CE}}\big(g(f(x_i)), y_i\big)} \Big)}{\sum_{x_j\in \mathcal{B}}\exp{\Big( \frac{1}{\lambda} l_{\text{CE}}\big(g(f(x_j)), y_j\big)} \Big)} $$

where $w_{x_i}$ is the re-weighting factor for the example $x_i$, $\mathcal{B}$ is the current batch, and $\lambda$ is a temperature parameter. The re-weighting factor is basically the softmax of the loss of the current batch.

For UDA~\cite{DBLP:conf/nips/XieDHL020UDA}, we leverage the augmented data in consistency training. In addition to the cross-entropy loss of the original data, we jointly train with the objective that minimizing the consistency loss between original data and augmented data:

\begin{align}
    \mathcal{L} &= \sum_{i=1}^n \Big( l_{\text{CE}}\big(g(f(x_i)), y_i\big) \\ \nonumber
     &\quad \quad + \alpha_c \sum_{j=1}^k  \text{KL}\big(\ g(f(x_i))\ ||\ g(f(x_{i,j}'))\ \big)  \Big)
\end{align}

where $x_{i, j}'$ is the $j$-th augmented example derived from $x_i$. $\alpha_c$ is the hyper-parameter to control the effect of consistency training. It's set as 10 after sufficient parameter searching.

\begin{table}[t]
\small
\centering
\renewcommand\arraystretch{1.1}
\setlength\tabcolsep{2.0pt}
\begin{tabular}{lcccccc}
\toprule
\multirow{2}{*}{Method} & \multicolumn{2}{c}{TREC} & \multicolumn{2}{c}{Irony} & \multicolumn{2}{c}{AGNews} \\ \cline{2-7}
& 1\% & 10\% & 1\% & 10\% & 0.05\% & 0.1\%  \\
\midrule
Back-Trans. (BT)&62.55 &93.62&52.29&64.69&85.39&86.35\\
BT+OD&62.19&94.67 &57.50&64.57&85.53&86.74\\
BT+OD+SR&\textbf{65.02}&\textbf{95.65}&\textbf{58.10}&\textbf{65.28}&\textbf{86.03}&\textbf{86.83}\\
\bottomrule
\end{tabular}
\caption{Experiments on using back-translation as the backbone data augementation method.}
\label{table:back_translation}
\end{table}

Besides using EDA as the backbone data augmentation method, we also test our ODDA framework on back-translation\footnote{We use the implementation from the nlpaug package (https://github.com/makcedward/nlpaug)} in \Cref{table:back_translation}. We can find that the ODDA framework can also work on back-translation, indicating a good generalizability of our framework.

\begin{table*}[t]
\small
\centering
\renewcommand\arraystretch{1.1}
\setlength\tabcolsep{2.0pt}
\begin{tabular}{lcccccccccc}
\toprule
\multirow{2}{*}{} & \multicolumn{2}{c}{TREC} & \multicolumn{2}{c}{Irony} & \multicolumn{2}{c}{AGNews} & \multicolumn{2}{c}{Sentiment} & \multicolumn{2}{c}{Offense} \\ \cline{2-11}
& 1\% & 10\% & 1\% & 10\% & 0.05\% & 0.1\% & 1\% & 10\% & 0.1\% & 1\% \\
\midrule
Optimizer & \multicolumn{10}{c}{AdamW} \\
Weight Decay & \multicolumn{10}{c}{1e-3}\\
Adam $\epsilon$ & \multicolumn{10}{c}{1e-8}\\
LR & \multicolumn{10}{c}{2e-5}\\
Batch Size & \multicolumn{10}{c}{32} \\
Max Length & \multicolumn{10}{c}{512} \\
Organic Epoch       & 40 & 30 & 100 & 20 & 30 & 30 & 30 & 10 & 30 & 30 \\
Augmentation Epoch  & 40 & 30 & 100 & 30 & 30 & 30 & 30 & 10 & 30 & 30 \\
Evaluation Interval & 1 & 5   & 1   & 1  & 5  & 5  & 5  & 20 & 1  & 5  \\
Temperature $\tau$ & 1 & 1 & 1 & 1 & 2 & 2 & 0.5 & 0.5 & 1 & 1 \\
SR $\alpha$ & 10 & 10 & 10 & 10 & 10 & 10 & 10 & 10 & 10 & 10 \\
\bottomrule
\end{tabular}
\caption{Hyperparameters for text classification experiments.}
\label{table:hparam_text_cls}
\vspace{-0.1in}
\end{table*}

\subsection{More Details about Question Answering} \label{sec:hparam_qa}

For question answering tasks, following previous works~\cite{DBLP:conf/aaai/SakaguchiBBC20, DBLP:conf/emnlp/YangMFSBWBCD20GDAUG}, we use RoBERTa as the base encoder.
For each question-option pair, the input format is then \texttt{[CLS] context [SEP] option [SEP]}. We take the embedding of the \texttt{[CLS]} token as the representation of the question-option pair. 
Then an MLP + softmax layer is put after the embeddings of the $c$ options, and the model is optimized with cross-entropy loss given a correct option.

WinoGrande is a commonsense reasoning benchmark to explore hard coreference resolutions problems such as ``The fist ate the worm, \_\_\_ was tasty'' (choose from ``fish'' and ``worm'').
It's hard as it requires commonsense knowledge that ``the subject of \textit{eat} tends to be hungry and the object of \textit{eat} tend to be tasty'', while machine learning models may associate ``fish'' with ``tasty'' with larger likelihood as they frequently co-occur in human corpora.
The WinoGrande dataset is composed of 5 subsets with different sizes, XS ($n=160$), S ($n=640$), M ($n=2558$), L ($n=10234$), and XL ($n=40398$).

CommonsenseQA is a commonsense question answering dataset constructed from the commonsense knowledge in ConceptNet~\cite{DBLP:conf/aaai/SpeerCH17}.
It aims to study the commonsense relations among daily entities within certain context.
For example, the correct answer to ``Where would you store a pillow case that is not in use?'' is ``drawer''. There are some distractor options such as ``bedroom'', which is a common place where a pillow locates without the context ``not in use''.

The augmentation method that we use for solving commonsense question answering is Generative Data Augmentation~\cite{DBLP:conf/emnlp/YangMFSBWBCD20GDAUG}. It uses three generation models to generate questions, correct answers, and distractors, respectively. Then in the data selection phase, influence function and a specifically designed heuristics that favors diverse synthetic data are used to select high-quality synthetic data.
Then the model is trained with a two-stage finetuning, where they first finetune the QA model on the synthetic data, and then finetune on the original data.
We use the released augmented data from~\citet{DBLP:conf/emnlp/YangMFSBWBCD20GDAUG}.
The number of augmented instances for each dataset is presented in Table~\ref{table:qa_stat}.
The hyperparameters that are used for the experiments for QA are presented in Table~\ref{table:qa_hyper_param}.

\section{Self-Regularization}\label{sec:appendix_sr}

We explain the reasons why Self-Regularization can serve as a denoising channel and yield better performance.
It is shown that the following fine-tuning method can enhance the robustness of representation learning, which provide guarantees for stochastic gradient descent algorithms by bounding some divergence between model at step $t$ and $t+1$~\cite{DBLP:journals/corr/abs-1301-3584}:

\begin{align}
\begin{aligned}
\text{arg min}&_{\Delta \theta}\ \mathcal{L}(\theta + \Delta \theta)    \\
& s.t.\ KL( f(\cdot, \theta_f) || f(\cdot, \theta_f + \Delta \theta_f) ) = \epsilon\\
\end{aligned}
\end{align}

Here, $f$ is a function that outputs vector representations, $\theta$ is the trainable parameters.
An approximation to this computationally intractable equation is proposed as follows~\cite{DBLP:conf/iclr/AghajanyanSGGZG21}:

\begin{align}
\small
    \begin{aligned}
    \mathcal{L}(f, g, \theta)& = \mathcal{L}(\theta) + \lambda KL_S(g\cdot f(x) || g\cdot f(x+z)) \\
    & s.t.\ z\sim \mathcal{N}(0, \sigma^2 I)\ \text{or}\ z\sim \mathcal{U}(-\sigma, \sigma)\\
    \end{aligned}
\end{align}

Here $g$ is a function that converts the output embedding of $f$ to a probability distribution. $KL_S$ is the symmetric KL divergence, and $z$ is sampled from the corresponding distribution as small perturbations. 
Instead of providing small perturbations using a random noise, Self-Regularization provide such perturbation with two different dropouts, which has shown to be effective in previous works~\cite{DBLP:conf/nips/LiangWLWMQCZL21}.

Moreover, there are other empirical findings that favors the effect of self-regularization in terms of denoising.
Noisy examples tend to be frequently forgotten after training for a long time~\cite{DBLP:conf/iclr/TonevaSCTBG19}, 
since the noise conflict with what have been learned in the model and the prediction can vary.
Self-regularization can be an alternative objective that mitigate the importance of the example.

\begin{table*}[t]
\small
\centering
\renewcommand\arraystretch{1.1}
\setlength{\tabcolsep}{2mm}{} 
\begin{tabular}{l|ccccc|p{5em}<{\centering}}
\toprule
  & \multicolumn{5}{c|}{WinoGrande} & \multirow{2}{*}{CSQA}\\
  & XS & S & M & L & XL & \\
\midrule
\# Original  & 160 & 640 & 2,558 & 10,234 & 40,398 & 9,727 \\
\# Synthetic  & 52,346 & 97,733 & 127,509 & 132,849 & 136,052 & 50,014 \\
\bottomrule
\end{tabular}
\caption{Number of training instances for WinoGrande and CommonsenseQA.} \label{table:qa_stat}
\vspace{-0.1in}
\end{table*}

\begin{table*}[t]
\small
\centering
\renewcommand\arraystretch{1.1}
\setlength{\tabcolsep}{2mm}{} 
\begin{tabular}{l|cccccp{5em}<{\centering}}
\toprule
  & \multicolumn{5}{c}{WinoGrande} & \multirow{2}{*}{CSQA}\\
  & XS & S & M & L & XL & \\
\midrule
Optimizer & \multicolumn{5}{c}{AdamW}  & AdamW \\
Weight Decay & \multicolumn{5}{c}{0.01} & 0.01 \\
Adam $\epsilon$ & \multicolumn{5}{c}{1e-6} & 1e-6 \\
LR synthetic & \multicolumn{5}{c}{5e-6} & 5e-6 \\
LR organic   & \multicolumn{5}{c}{1e-5} & 1e-5 \\
Batch Size & \multicolumn{5}{c}{16}  & 16 \\
Max Length & \multicolumn{5}{c}{70}  & 70 \\
Synthetic Epoch & 1 & 1 & 1 & 1 & 1 & 1 \\
Organic Epoch & 10 & 8 & 5 & 5 & 5  & 5 \\
LR Decay & \multicolumn{5}{c}{Linear}  & Linear\\
Warmup Ratio & \multicolumn{5}{c}{0.06}  & 0.06 \\
SR Warmup Steps & 2000 & 5000 & 5000 & 7000 & 7000 & 2500  \\
$\tau$ &    2  & 1   & 1   & 1 & 1     &  1 \\
$\alpha$ & 0.5 & 0.1 & 1.0 & 0.5 & 0.5 & 0.5  \\
\bottomrule
\end{tabular}
\caption{Essential Hyperparameters for WinoGrande and CommonsenseQA.} \label{table:qa_hyper_param}
\vspace{-0.1in}
\end{table*}

\end{document}